\documentclass[journal]{IEEEtran}
\ifCLASSINFOpdf
\else
\fi
\usepackage{graphicx}

\usepackage{amsmath,amssymb}
\usepackage{amsfonts}

\usepackage{tikz}
\usepackage{textcomp}
\usepackage{hyperref}
\usepackage{lipsum}

\newcommand\copyrighttext{%
	\footnotesize \textcopyright © 2018 IEEE. Personal use of this material is permitted. Permission from IEEE must be obtained for all other uses, in any current or future media, including reprinting/republishing this material for advertising or promotional purposes, creating new collective works, for resale or redistribution to servers or lists, or reuse of any copyrighted component of this work in other works.
}

\newcommand\copyrightnotice{%
	\begin{tikzpicture}[remember picture,overlay]
	\node[anchor=south,yshift=3pt] at (current page.south) {\fbox{\parbox{\dimexpr\textwidth-\fboxsep-\fboxrule\relax}{\copyrighttext}}};
	\end{tikzpicture}%
}

\begin{document}
%
\title{Avoiding overfitting of multilayer perceptrons by training derivatives}
%
%
%

\author{V.I. Avrutskiy
	\thanks{V.I. Avrutskiy is with the Department of Aeromechanics and Flight Engineering of Moscow Institute of Physics and Technology, Institutsky lane 9, Dolgoprudny, Moscow region, 141700, e-mail: avrutsky@phystech.edu}
}

%
%

\markboth{Submitted to IEEE Transactions on Neural Networks and Learning Systems}%
{Shell \MakeLowercase{\textit{et al.}}: Bare Demo of IEEEtran.cls for IEEE Journals}
%



\maketitle

\copyrightnotice
\begin{abstract}
Resistance to overfitting is observed for neural networks trained
with extended backpropagation algorithm. In addition to target values,
its cost function uses derivatives of those up to the 4\textsuperscript{th} order.
For common applications of neural networks, high order derivatives are
not readily available, so simpler cases are considered: training
network to approximate analytical function inside 2D and 5D domains and solving Poisson equation inside a 2D circle. For function approximation, the cost is a sum of squared differences between output and target as well as their derivatives with respect to the input. Differential equations are usually solved by putting a multilayer perceptron in place of unknown function and training its weights, so that equation holds within some margin of
error. Commonly used cost is the equation's residual squared. Added terms are squared derivatives of said residual with respect to the independent variables. To investigate overfitting, the cost is minimized for points of regular grids with various spacing, and its root mean is compared with its value on much denser test set. Fully connected perceptrons with six hidden layers and $2\cdot10^{4}$, $1\cdot10^{6}$ and $5\cdot10^{6}$ weights in total are trained with Rprop until cost changes by less than 10\% for last 1000 epochs, or when the 10000\textsuperscript{th} epoch is reached.
Training the network with $5\cdot10^{6}$ weights to represent simple 2D function using 10 points with 8 extra derivatives in each produces cost test to train ratio of $1.5$, whereas for classical backpropagation in comparable conditions this ratio is $2\cdot10^{4}$.
\end{abstract}


%
\IEEEpeerreviewmaketitle

\section{Introduction}
%
%
%
%
\IEEEPARstart{O}{verfitting} is a common fault of neural networks that occurs in many
different applications  \cite{tetko1995neural,lawrence1997lessons,krizhevsky2012imagenet,lawrence2000overfitting}. Consider a case of supervised learning: network
$\mathcal{N}$ is to be trained to represent vector function $F:X\rightarrow Y$ known
only by a finite set of its arguments $\mathbf{x}_{\mathrm{a}}\subset X$ and corresponding
outputs $\mathbf{y}_{\mathrm{a}}\subset Y$. It is a common practice to subdivide $\mathbf{x}_{\mathrm{a}}$,$\mathbf{y}_{\mathrm{a}}$
into at least two sets: training $\mathbf{x}_{\mathrm{b}}$,$\mathbf{y}_{\mathrm{b}}$ and test $\mathbf{x}_{\mathrm{c}}$,$\mathbf{y}_{\mathrm{c}}$.
Whatever algorithm is used for minimization, it is not run for elements
of $\mathbf{x}_{\mathrm{c}}$,$\mathbf{y}_{\mathrm{c}}$. This is done to later observe network's performance
on previously unknown data. Overfitting can be broadly described as
different behavior of cost function $E=||\mathcal{N}(\vec{x})-\vec{y}||$ for
$\vec{x}\in\mathbf{x}_{\mathrm{b}}$, $\vec{y}\in\mathbf{y}_{\mathrm{b}}$ and $\vec{x}\in\mathbf{x}_{\mathrm{c}}$, $\vec{y}\in\mathbf{y}_{\mathrm{c}}$. For example, its average values on those two sets can be orders of magnitude away. In broad terms, the ratio between cost function averaged on test and train set shows how much attention training procedure pays to the actual input rather than to the function behind it. However, this relation between averages is not important all by itself, since the absolute values of $E_{\mathrm{c}}$ on test set $\mathbf{x}_{\mathrm{c}}$,$\mathbf{y}_{\mathrm{c}}$ are the goal of training procedure.

There are numerous ways to tackle with overfitting. It is possible to directly address negative effects for $E_{c}$ by introducing a special set of patterns $\mathbf{x}_{\mathrm{cv}}$,$\mathbf{y}_{\mathrm{cv}}$ 
called cross validation set. Its elements are not fed to minimization
algorithm as well, but during the training, cost function for them is observed.
When it starts to increase, while the cost on $\mathbf{x}_{\mathrm{b}}$,$\mathbf{y}_{\mathrm{b}}$ continues
to lower, one can draw a conclusion that negative effect of overfitting
started to overpower positive effects of the training itself. If one
choses to stop, the decision effectively puts a lower bond on approximation
error. One might also say that before $E_{\mathrm{c}}$ started to increase,
any presence of overfitting was not practically significant. Many ways to
handle such situations were developed \cite{sarle1996stopped,lecun1990optimal,hassibi1993optimal}. For example, neural network
can be pruned \cite{reed1993pruning}, or some special statistical rule for updating weights
can be used \cite{hinton2012improving,srivastava2014dropout,zaremba2014recurrent}. Generally, the more approximation abilities a neural
network has the more it is prone to overfitting. 

If random noise is present and the network is overtrained, i.e. backpropagation algorithm was run long enough to start paying too much attention to the training set, it almost inevitably leads to worsening of performance on the test set, since noise on the one is not correlated with noise on the other. However, if data is noiseless, then even having an extremely low error on training points can bring no negative effects to cost on the test set. This paper is using various derivatives of target function and, therefore, is unable to take noise into consideration, since even the slightest variations of neighboring points can have profound consequences to numerically calculated derivatives even of the first, not to mention higher orders. Vice versa, if derivatives up to say the 4\textsuperscript{th} order are known then nearby points can be calculated by Taylor approximation virtually without any noise. Therefore, in the scope of this paper overfitting is defined not as a worsening of performance on the test set, which was not observed, but rather as an ability of network to pay more attention to the training set. Thus, an absence of overfitting can be expressed as an inability to distinguish between statistical properties of the cost function on the test and training sets.

\section{Extended backpropagation}
Algorithms for supervised training of neural networks with cost functions that include
derivatives were described in a number of publications \cite{flake2000differentiating,lagaris1997artificial}. Most
notable are double backpropagation \cite{drucker1992improving} and tangent propagation \cite{simard1998transformation}.
They both consider image classification problem and their cost functions
in addition to squared difference between output and target classes include
the first order derivatives of that difference. Double backpropagation calculates
them with respect to values of individual input pixels, so derivatives
of output can be viewed as a slope of network's reaction to subtle
distortions. Derivatives of target with respect to distortions are
zero since they do not change class, thus, training tries to minimize
derivatives of output with respect to input for each meaningful pattern.
Tangent propagation is more elaborate and calculates derivatives along
special directions of input space. For each image $I$, a direction
is another pseudo-image $J$ that can be multiplied by small factor $\alpha$
and added to $I$, so that $I+\alpha J$ simulates an infinitesimal geometrical
distortion of $I$, for example, a rotation through small angle. Such
distortion do not change class either, so derivative of output with respect to $\alpha$ is trained
to be zero. Described methods are rarely used in practice, which might
be due to their limited benefits, an increase in training time and a tricky
process of calculating derivatives. This study does not consider complicated
mappings like image classifiers, and tasks are limited to somewhat
artificial, where functions are from space ${\mathbb{C}}^{\infty}$, and
their derivatives are easily computable. This allows to closely investigate
effects of using high order terms on perceptron training.

Another case when derivatives appear in cost function, is solving
differential equations \cite{kumar2011multilayer,lagaris1998artificial,malek2006numerical,shirvany2009multilayer}. Classical approach is to construct a
neural network that maps each vector of independent variables to solution.
Consider a boundary value problem inside a region $\Gamma$ for function
$u(x,y)$ written as:
\[
U(x,y,u,u_{x},u_{y},u_{xx},...)=0
\] 
\[
\left.u\right|_{\partial\Gamma}=f
\]
The first step is to make a substitution using relation:
\begin{equation}\label{zamena}
u(x,y)=v(x,y)\cdot\phi(x,y)+f
\end{equation}
where $\phi(x,y)$ is a known function that vanishes on the boundary $\partial\Gamma$ and is non zero inside $\Gamma$, and $v(x,y)$ is to be constructed by a neural network \cite{lagaris1998artificial}. The term $f$ is now a smooth continuation of the boundary condition into $\Gamma$ that is supposed to exist. The equation is written as:
\[
V(x,y,v,v_{x},v_{y},v_{xx},...)=0
\]
and the boundary condition is simply:
\[\left.v\right|_{\partial\Gamma}<\infty\]
A network with two inputs, one output and a suitable number of hidden layers and neurons is then created. Its weights are initialized and trained to minimize cost function $E$ - a measure of a local cost function $e=V^{2}$ on $\Gamma$ calculated using a finite set of grid points. If it reaches small enough values, ${v(x,y)\cdot\phi(x,y)+f}$ can be considered as an approximate numerical solution, and as long as the procedure converges to a finite $v$, the boundary condition is satisfied. An extension of classical backpropagation is required due to the presence in $e$ terms like $v_{xx}$, which are derivatives of the neural network's output with respect to input. Various types of differential equations can be solved using this approach \cite{lagaris1997artificial,zuniga2017solving}. The smaller effect of overfitting is, the fewer points are required to capture the behavior of function $e$ on $\Gamma$ for proper discretization of $E$ and, therefore, to solve the equation in the entire region.

Study \cite{avrutskiy2017enhancing} found significant improvements from using high order derivatives in learning process. For classical supervised training, the cost, which is equal to squared difference between output and target, was added with squared differences of their derivatives with respect to input up to the 4\textsuperscript{th} order. In the study target function was chosen as a piece of 2D Fourier series and all derivatives were readily available. Such modifications were found to enhance precision up to $\sim100$ times and allowed to use three times less points for training, but required to increase the number of neurons and layers as well. Another obvious cost of such enhancements was that derivatives of target must be known beforehand. For the case of solving partial differential equations the situation is different. The expression $V$ behind the cost function $e$ is simply all terms of the equation put together, therefore, it is possible to introduce extra derivatives simply by applying operators like $\partial/\partial x$, $\partial/\partial y$ to $V$ and adding squares of the results to the local cost function. For partial differential equations inclusion of extra derivatives up to the second order allowed to use 8 times less grid points and obtain 5 times smaller error.
Paper \cite{avrutskiy2017enhancing}, however, was mostly focused on precision and implemented a cascade procedure when training is started with all available derivatives, and then higher order terms are turned off as the process continues. A contradiction between the necessity to increase network size and decrease in minimum number of points that prevent overfitting leads to this study. Since absolute precision is not the goal, the process of altering cost functions is omitted and instead the one with maximum number of available derivatives is used. Overfitting properties of extended training are investigated for both cases - supervised learning and solving PDE.

\section{Method}
\subsection{Direct function approximation}
Neural network is trained to represent a scalar function ${f:\mathbb{R}^{n}\rightarrow\mathbb{R}^{1}}$ defined inside a region $\Gamma\in\mathbb{R}^{n}$. Components of the input vector $\vec{x}$ are denoted by $x_{i}$. Network's output is denoted by $\mathcal{N}=\mathcal{N}(\vec{x})=\mathcal{N}(x_{1},...,x_{n})$.
To gather a cost function the following terms are used:
\begin{equation}\label{nevazkif}
e_{k}=\sum_{i}\left[\frac{\partial^{k}}{\partial x_{i}^{k}}\mathcal{N}-\frac{\partial^{k}}{\partial x_{i}^{k}}f\right]^{2}
\end{equation}
Here $i$ can run through all or some of the input variables. If $k=0$, all terms are the same and the sum can be omitted turning the expression into classical backpropagation cost: ${(\mathcal{N}-f)^{2}}$. The next order term $k=1$ is used in double backpropagation. The cost of order $s$ for one pattern (i.e. local) is defined as follows:
\begin{equation}\label{totallocf}
e_{s}^{\mathrm{local}}=\sum_{k=0}^{s}e_{k}
\end{equation}
A network is trained on a set of $M$ patterns, so the total cost is:
\[
E_{s}=\frac{1}{M}\sum_{\alpha=1}^{M}e_{s}^{\mathrm{local}}(\vec{x}_{\alpha})
\]
Results for $s=0$ will be compared with the ones for $s=4$. 
\subsection{Solving differential equation}
Boundary value problem inside 2-dimensional region $\Gamma$ is considered: 
\[
U(x_{1},x_{2},u,u_{x_{1}},u_{x_{1}x_{1}},...)=0
\]
\[
\left.u\right|_{\partial\Gamma}=f
\]
According to \eqref{zamena} the function is substituted, and the equation is written for $v$:
\[
V(x_{1},x_{2},v,v_{x_{1}},v_{x_{1}x_{1}}...)=0
\]
\[
\left.v\right|_{\partial\Gamma}<\infty
\]
which is to be represented by a neural network $v=\mathcal{N}$. Boundary condition is satisfied automatically. Similarly to \eqref{nevazkif} terms of different order are written as:
\[
e_{k}=\sum_{i=1}^{2}\left[\frac{\partial^{k}}{\partial x_{i}^{k}}V\right]^{2}
\]
And one can omit the sum for $k=0$. Classical method of solving PDE involves only $e_{0}=V^{2}$. The local cost function of the order $s$ is similar:
\begin{equation}\label{deqloce}
e_{s}^{\mathrm{local}}=\sum_{k=0}^{s}e_{k}
\end{equation}
Used grids are very close to regular, therefore, the total cost function which is a measure of $e_{s}^{\mathrm{local}}$ on $\Gamma$ can be discretized on $M$ points as:
\[
E_{s}=\underset{\vec{x}\in\Gamma}{\mu}(e_{s}^{\mathrm{local}})\simeq\frac{1}{M}\sum_{\alpha=1}^{M}e^{\mathrm{local}}_{s}(\vec{x}_{\alpha})
\]
Results for $s=0$ will be compared with the ones for $s=4$.
\section{Results}
Analytical functions are approximated inside 2D box and 5D unit sphere. Boundary value problem is solved inside 2D unit circle. Regions are denoted as $\Gamma$ and their boundaries as $\partial\Gamma$. For all cases the cost is minimized on grids that are generated in similar manner and comprised of two parts - internal and surface. The internal part is a regular grid, which is generated in three steps. At first, a Cartesian grid with spacing $\lambda$ inside sufficient volume is created. Two random vectors in input space are then chosen, and the grid is rotated from one to another. Finally, it is shifted along each direction by a random value from interval $[-\frac{\lambda}{4},\frac{\lambda}{4}]$, and after that all points outside of $\Gamma$ are thrown away. The surface part contains equidistant points from $\partial\Gamma$ with distance~$\tau=\lambda$ unless otherwise stated.

Training is based on RProp \cite{riedmiller1993direct} procedure with parameters $\eta_{+}=1.2$, $\eta_{-}=0.5$.
Weights are forced to stay in $[-20,20]$ interval, no min/max bonds
for steps are imposed. Initial step $\Delta_{0}$ is set to $2\cdot10^{-4}$. Due to the absence of a minimal step a lot of them are reduced to zero after certain number of epochs. To tackle that, after each 1000 epochs all zero steps are restored to $10^{-6}$. The precision is single. Weights are initialized \cite{he2015delving} with random values from range $\pm2/\sqrt{s}$, where $s$ is a number of senders. Thresholds are initialized in range $\pm0.1$. All of the layers are fully connected and have non linear sigmoid activation function:
\[\sigma(x)=\frac{1}{1+\exp(-x)}\]
except for the input and output layers which are linear. To make sure that backpropagation is run long enough to produce overfitting, weak rules for stopping are implemented. During both classical and extended training modes, the root mean of $e_{0}$ for grid points $\sqrt{\left\langle e_{0}\right\rangle_{\mathrm{train}}}$ is tracked and after its best value has changed by less than 10\% for the last 1000 epochs, or when the 10000\textsuperscript{th} epoch is reached, the training is stopped. The weights from the best iteration are saved, and for them $\sqrt{\left\langle e_{0}\right\rangle_{\mathrm{test}} }$ is calculated on a much finer grid. All results are linked not to the number of grid points, but to the number of parameters $N$ that were used for training. The value $N$ coincides with the number of grid points $M$, when only values of function are used in the cost, and is equal to $(p+1)M$, if $p$ extra derivatives in each point are trained. Plotting results against $N$ also equalizes the total number of arithmetic operations per one epoch provided networks are of the same architecture. For each training grid, method and network the process is repeated 5 times, and the average values are presented. Space above and below each plot is filled according to the maximum and minimum values obtained in those 5 runs.

\subsection{Direct function approximation}
\subsubsection{2D function}
\begin{figure}[!t]
	\centering
	\includegraphics[width=3.5in]{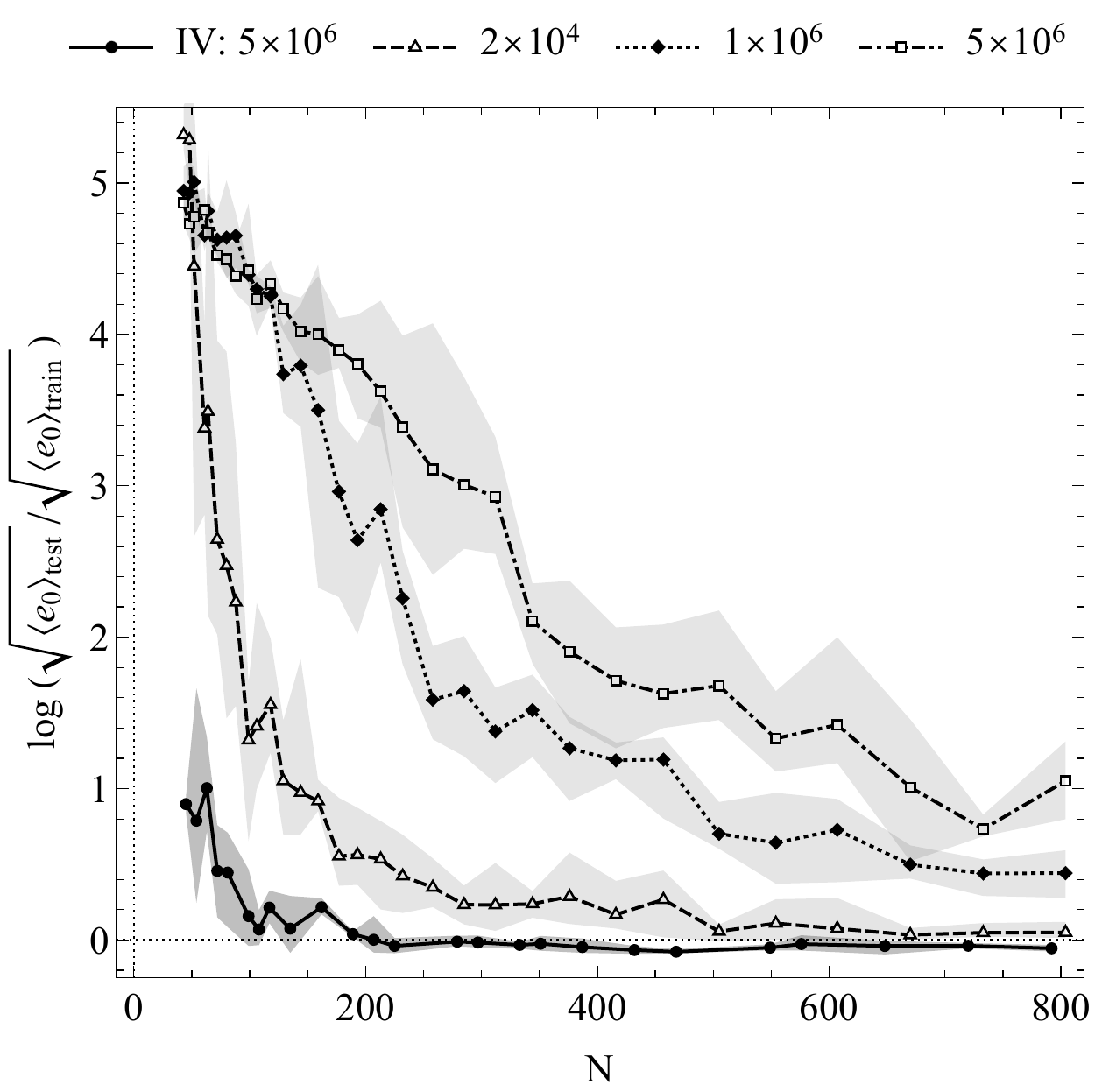}
	\caption{Approximating 2D function. Base 10 logarithm of the ratio between root mean square errors on test and train sets is plotted against the number of parameters used for training. Solid line is for the network with $5\cdot10^{6}$ weights trained by extended algorithm, and the rest curves are for networks trained by classical algorithm.}
	\label{fig:2d1}
\end{figure}

\begin{figure}[!t]
	\centering
	\includegraphics[width=3.5in]{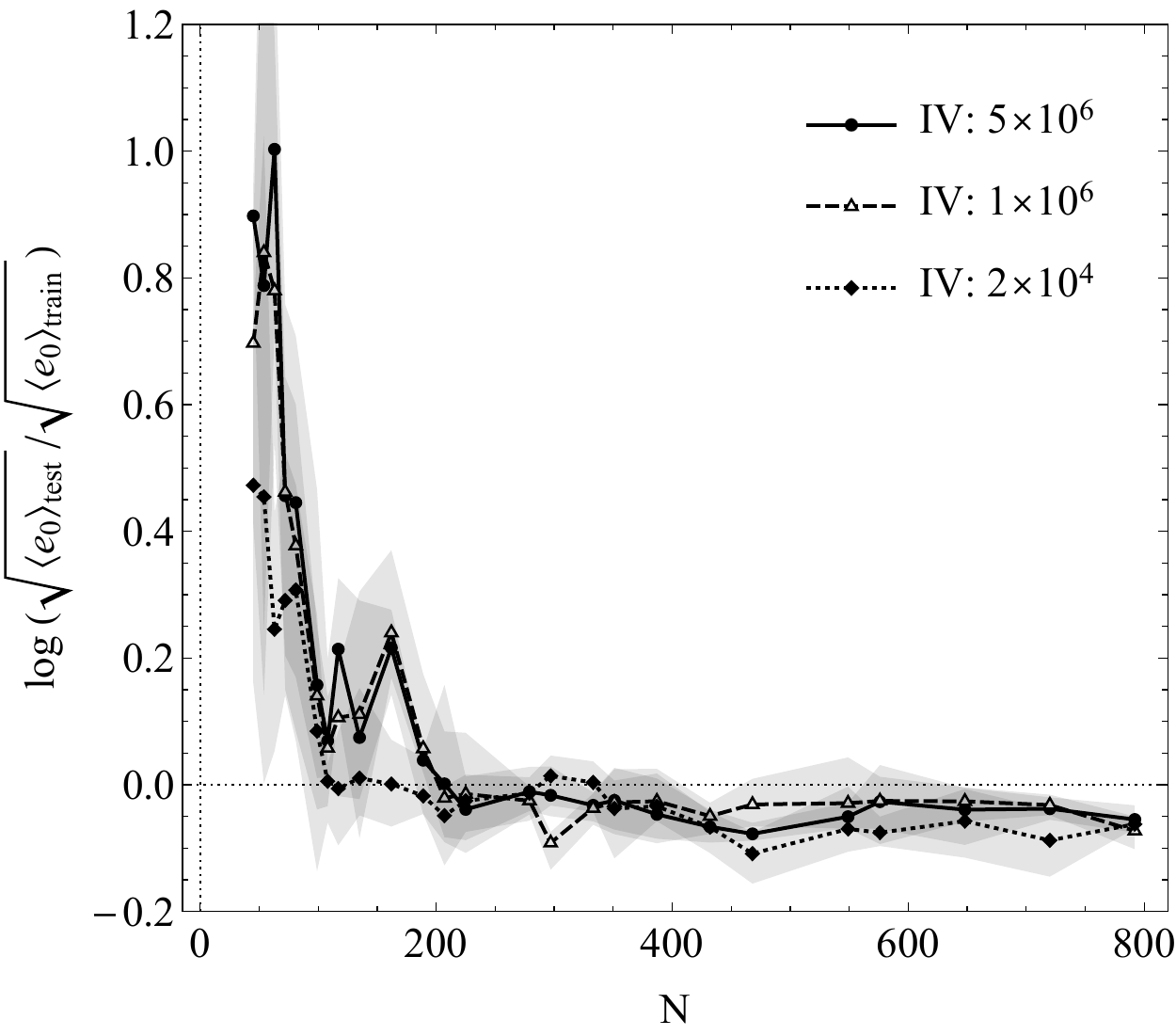}
	\caption{Approximating 2D function. Base 10 logarithm of the ratio between root mean square errors on test and train sets is plotted against the number of grid parameters. Comparison between networks with different capacities trained by extended algorithm, solid line is the same as on fig. \ref{fig:2d1}.}
	\label{fig:2d2}
\end{figure}

\begin{figure}[!t]
	\centering
	\includegraphics[width=3.5in]{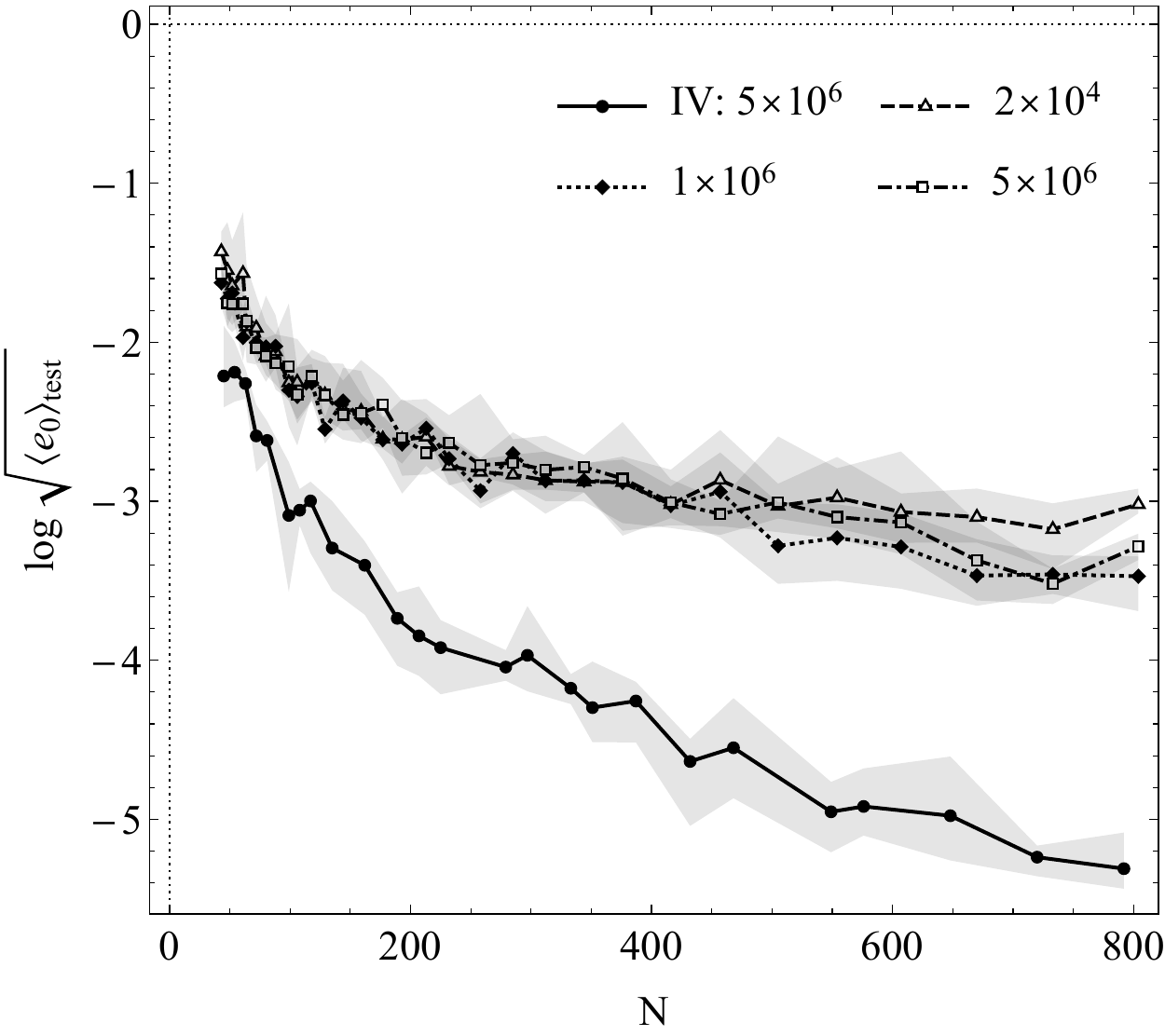}
	\caption{Approximating 2D function. Base 10 logarithm of root mean square error on the test set. The solid line is for extended algorithm, which utilizes derivatives up to the IV order.}
	\label{fig:2d3}
\end{figure}

networks are trained to represent the following expression inside $[-1,1]^{2}$ box:
\[
f=\frac{1}{2}\left(-x_{1}^{2}-x_{2}^{2}+1\right)+2\tanh\left(x_{1}\cdot\sin\frac{x_{2}}{2}\right)-\sin x_{1}\cdot\cos x_{2}
\]
A series of grids with various densities is generated. It starts with $\lambda=0.073$ and a grid containing 804 points. For each next grid $\lambda$ is increased, so that the total number of points would be approximately 10\% less. The last grid has $\lambda=1.45$ and 5 points, 4 on the boundary and 1 near the middle. In each point derivatives from the 1\textsuperscript{st} to the 4\textsuperscript{th} order are calculated with respect to $x_{1}$ and $x_{2}$. Extended training is using $i={1,2}$ in expression \eqref{nevazkif} and $s=4$ in \eqref{totallocf}, therefore, the total number of parameters for a grid of $M$ points is $N=9M$. For classical training $s=0$ and $N=M$. To plot curves on the same interval of $N$ extended training is run only on grids from 88 to 5 points ($\lambda$ from 0.24 to 1.45), and classical one is run on grids from 804 to 43 points ($\lambda$ from 0.073 to 0.37). It is worth mentioning that calculating the target function $9M$ times is exactly enough to evaluate derivatives up to the 4\textsuperscript{th} order with respect to $x_{1}$ and $x_{2}$ in $M$ points using the first order finite difference stencil. To investigate how network's architecture affects overfitting three different layer configurations are used:
\[
2,64,64,64,64,64,64,1
\]
\[
2,512,512,512,512,512,512,1
\]
\[
2,1024,1024,1024,1024,1024,1024,1
\]
They comprise $2\cdot10^{4}$, $1\cdot10^{6}$ and $5\cdot10^{6}$ weights respectively and are referred to on figures by those numbers. After training is finished, the performance of each network is tested on a grid with 3400 points and $\lambda=0.035$. Base 10 logarithm of the ratio $\sqrt{\left\langle e_{0}\right\rangle_{\mathrm{test}}}/\sqrt{\left\langle e_{0}\right\rangle_{\mathrm{train}} }$ is plotted on fig. \ref{fig:2d1}. The network with the most number of weights trained by extended algorithm is compared with three networks trained with classical cost function. The ratio itself and its variance, which is related to amount of filling above and below each curve, are much smaller for extended method.
Fig. \ref{fig:2d2} shows overfitting ratio for networks trained by extended algorithm. Plots for different architectures are very close to each other, unlike when the same networks are trained with classical cost function. From about 200 parameters and more, which corresponds to 23 grid points, all curves lie mostly below the zero, which means that the root mean square error calculated on the dense test grid is a bit smaller than the one calculated on the training grid.
Fig. \ref{fig:2d3} depicts the final root mean square error on the test set. It represents an inevitably more important aspect in a sense that the previous plots alone do not necessary mean that training was successful. Only one curve is shown for extended algorithm, since others lie very close to it.
\subsubsection{5D function}
networks are trained to represent the following expression inside 5D unit sphere $\Gamma:r\leq1$
\[
f=\frac{1}{2}(-x_{2}^{2}-x_{5}^{2}+1)+2\tanh\left(x_{3}\cdot\sin\frac{x_{4}}{2}\right)-\sin x_{1}\cos x_{2}
\]
A series of grids is generated. Unlike 2D case, $\tau=1.6\lambda$. The first grid has $\lambda=0.336$ and 1579 points. Each next grid has about 10\% less points. The last grid has $\lambda=1.1$ and 11 points, 10 of them are on the boundary and 1 near the middle. In each point derivatives from the 1\textsuperscript{st} to the 4\textsuperscript{th} order are calculated with respect to two randomly chosen directions $x_{p}$ and $x_{q}$ out of five possible. Extended training is using $i={p,q}$ in expression \eqref{nevazkif} and $s=4$ in \eqref{totallocf}, therefore the total number of parameters for a grid of $M$ points is $N=9M$. Extended training is run on grids from 166 to 11 points ($\lambda$ from 0.55 to 1.1) and classical training on grids from 1579 to 98 points ($\lambda$ from 0.336 to 0.62). Calculating target function $9M$ times is again exactly enough to evaluate derivatives with respect to two mentioned variables. Test grid has 11000 points and $\lambda=0.15$. Network configurations are the same as for 2D case, except now input layers have 5 neurons instead of 2.

\begin{figure}[!t]
	\centering
	\includegraphics[width=3.5in]{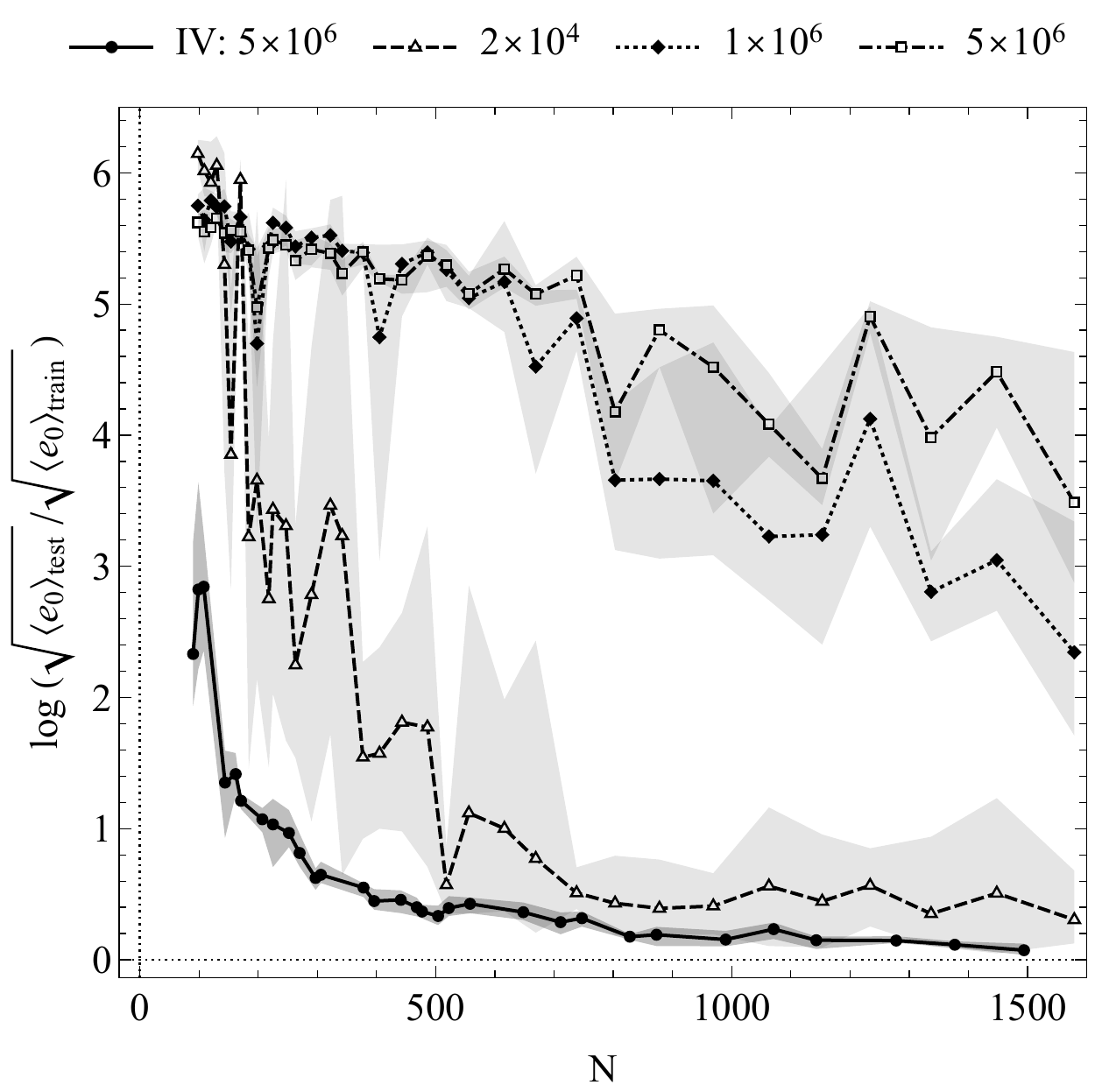}
	\caption{Approximating 5D function. Base 10 logarithm of the ratio between root mean square errors on test and train sets is plotted against the total number of parameters used for training. Solid line is for extended cost function, and the rest are for classical.}
	\label{fig:5d1}
\end{figure}

\begin{figure}[!t]
	\centering
	\includegraphics[width=3.5in]{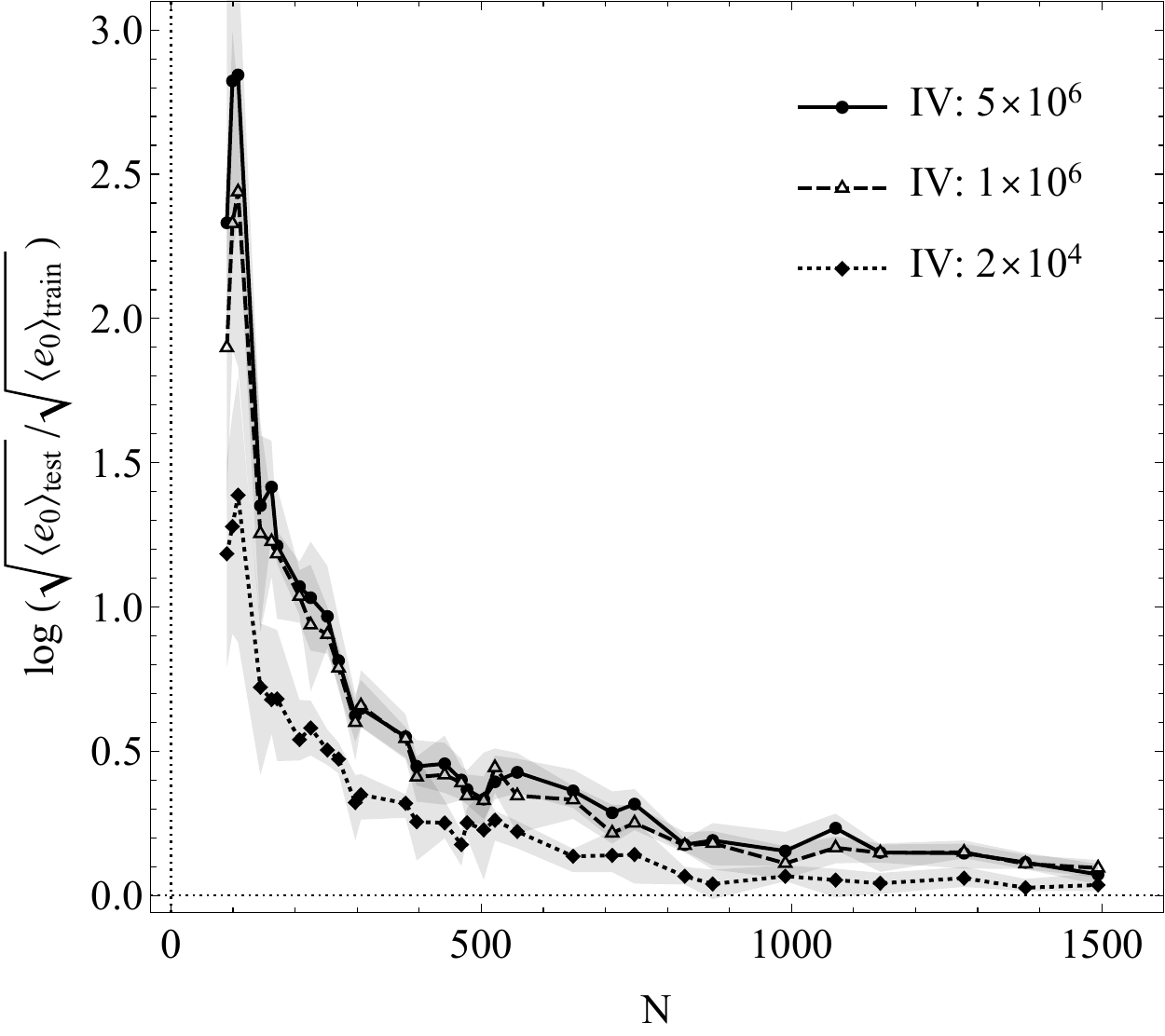}
	\caption{Approximating 5D function. Base 10 logarithm of the ratio between root mean square errors on test and train sets is plotted against the total number of grid parameters. Comparison between networks with different capacities trained with extended algorithm, solid line is the same as on fig. \ref{fig:5d1}.}
	\label{fig:5d2}
\end{figure}

\begin{figure}[!t]
	\centering
	\includegraphics[width=3.5in]{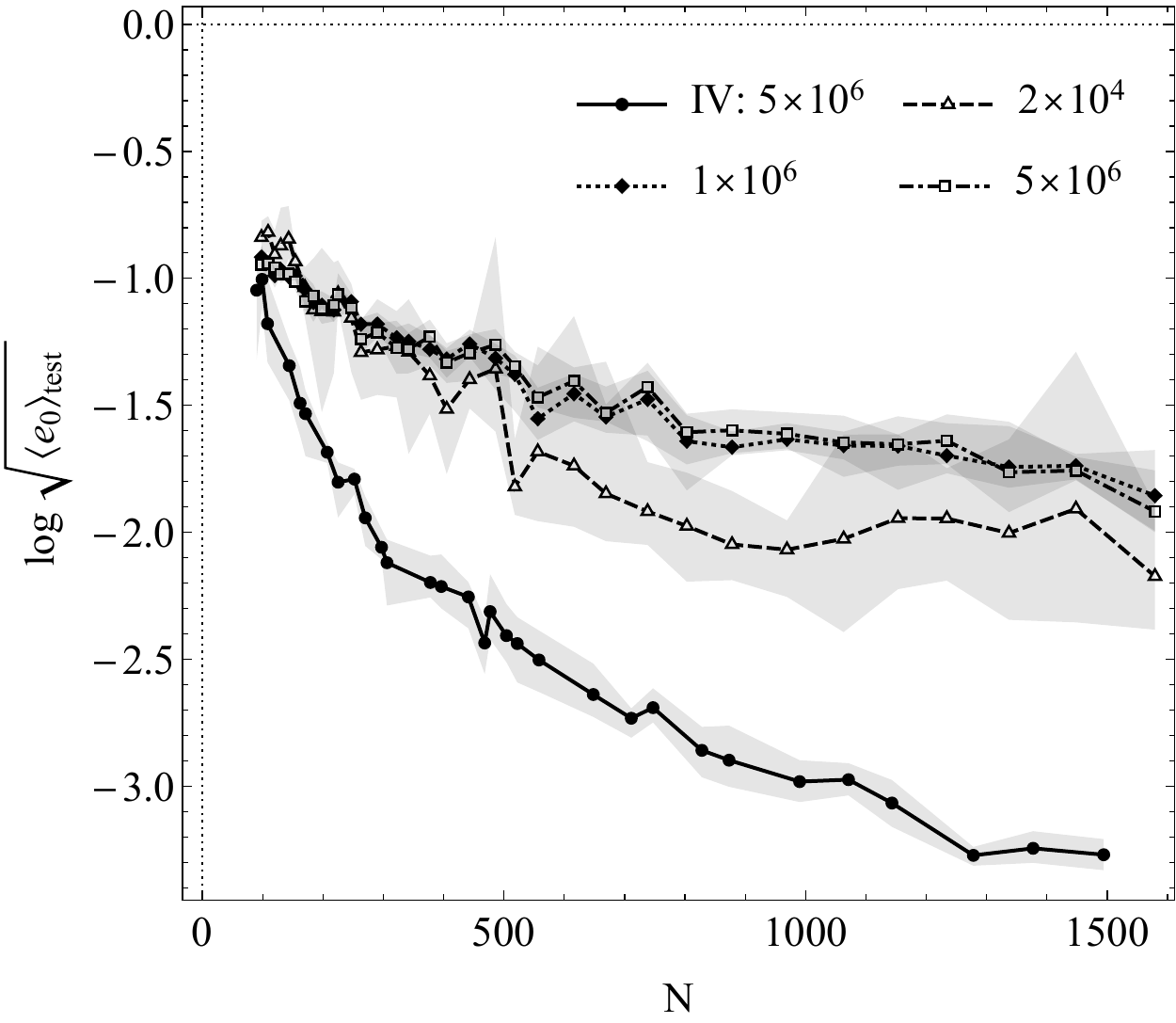}
	\caption{Approximating 5D function. Base 10 logarithm of root mean square error on the test set. Solid line is for extended algorithm, which utilizes derivatives up to the IV order.}
	\label{fig:5d3}
\end{figure}

Similarly to previous case, fig. \ref{fig:5d1} shows overfitting ratio for the biggest network trained with extended mode, and all networks trained with classical mode. Unlike 2D case, even on the most dense grids overfitting is very strong for larger networks trained without derivatives. Fig. \ref{fig:5d2} plots the same ratio for three architectures trained with extended cost. It is again small, barely different and has low variance between training attempts. Finally, fig. \ref{fig:5d3} compares performance of networks on the test set. It shows that extended mode not only produces very close test and train errors, but also a better approximation. Different curves for extended training are not shown since they lie very close to each other. One can notice that among results of classical training, the $1\cdot10^{4}$ network (the line marked with triangles) shows both: lower overfitting on fig. \ref{fig:5d1} and better test cost on fig. \ref{fig:5d3}. This could make the network a preferable choice for the task. However, when derivatives are used, any architecture demonstrates much smaller overfitting ratio and nearly one extra order in final precision. The only advantage of smaller network left is training time. Despite this being a 5D task, for each input vector it is enough to train derivatives with respect to two directions, provided they are randomly chosen from point to point.
\subsection{Solving differential equation}
\begin{figure}[!t]
	\centering
	\includegraphics[width=3.5in]{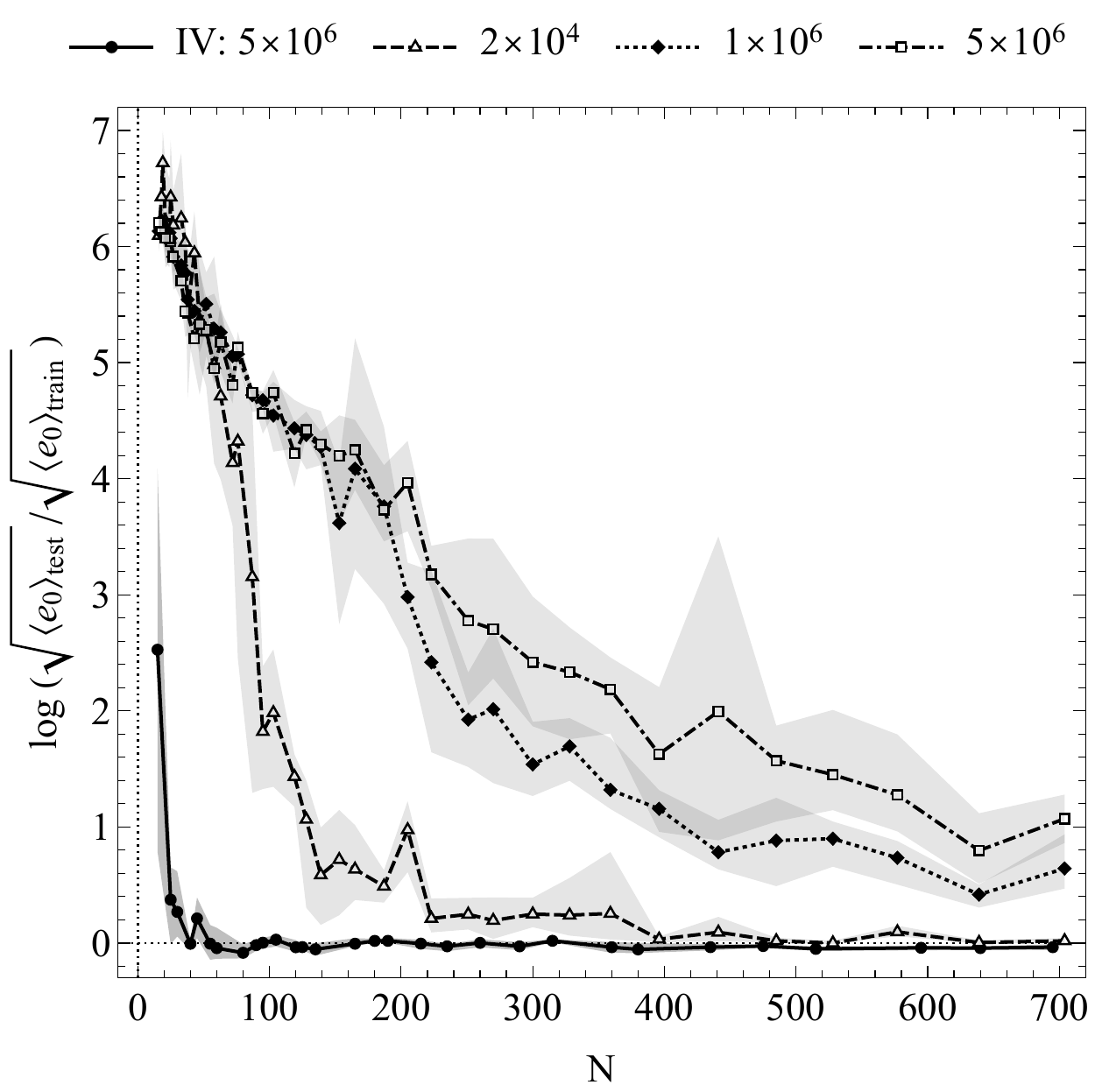}
	\caption{2D Boundary value problem. Base 10 logarithm of the ratio between root mean square of residuals on test and train grids is plotted against the equivalent number of grid points. Solid line is for extended cost function and the rest are for classical.}
	\label{fig:deq1}
\end{figure}

\begin{figure}[!t]
	\centering
	\includegraphics[width=3.5in]{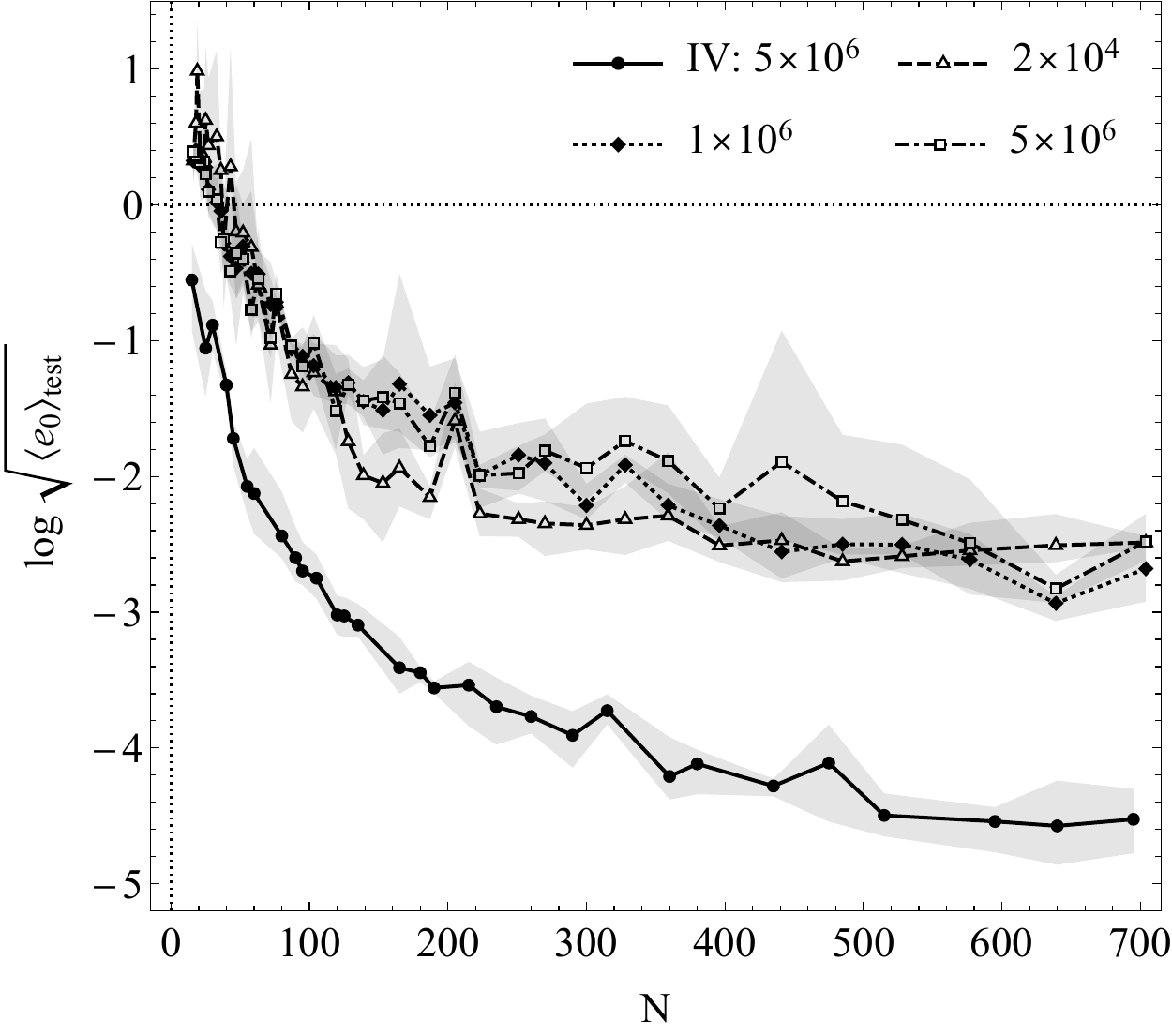}
	\caption{2D Boundary value problem. Base 10 logarithm of root mean square residual on the test set plotted against the equivalent number of grid points. Solid line is for extended algorithm, which utilizes derivatives up to the IV order.}
	\label{fig:deq2}
\end{figure}

\begin{figure}[!t]
	\centering
	\includegraphics[width=3.5in]{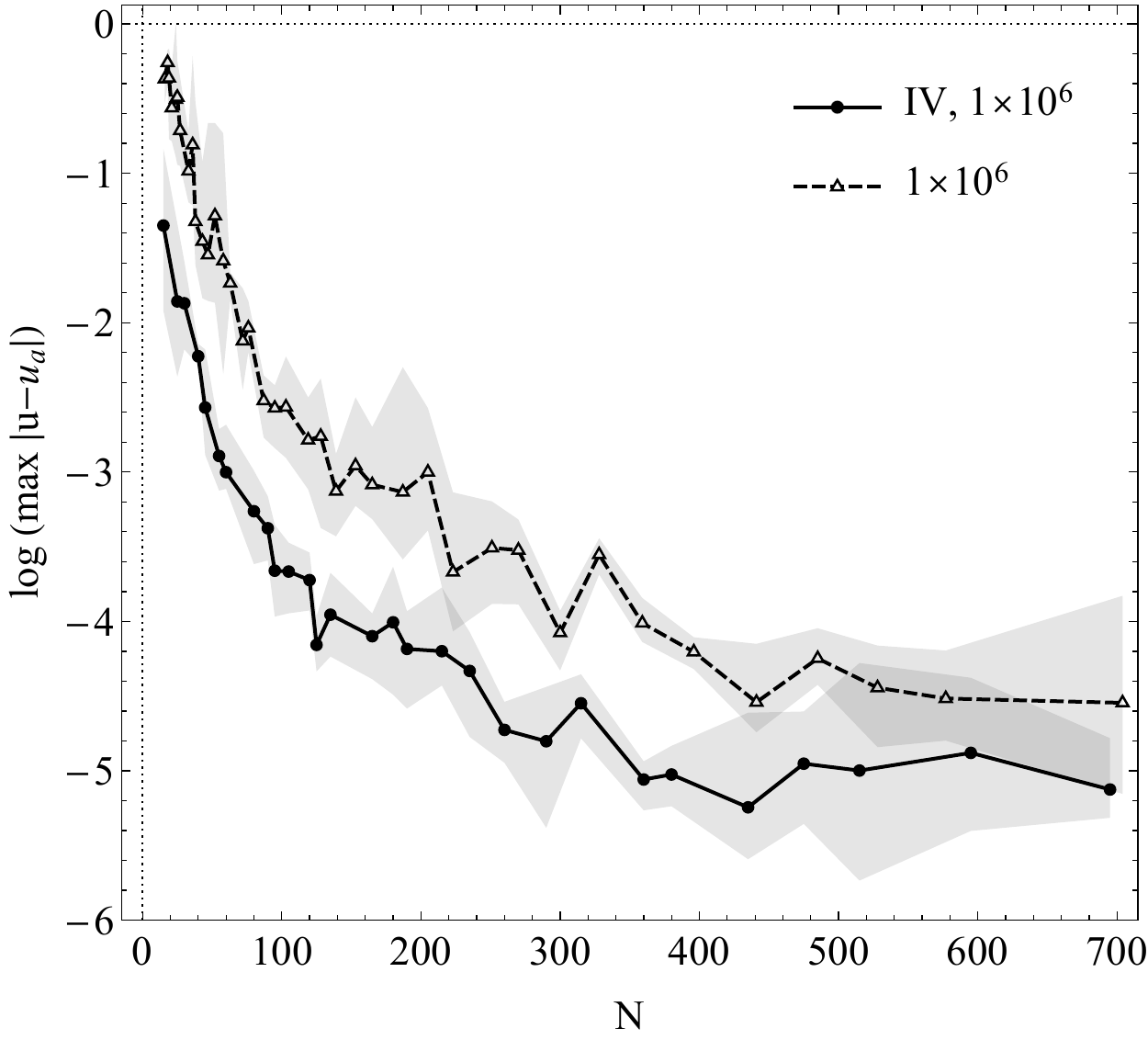}
	\caption{2D Boundary value problem. Base 10 logarithm of the maximum deviation from analytical solution calculated on the test set, $N$ is an equivalent number of grid points. Solid line is for extended algorithm, which utilizes derivatives of the residual up to the IV order.}
	\label{fig:deq4}
\end{figure}

Poisson equation inside 2D unit circle $\Gamma:x_{1}^{2}+x_{2}^{2}\leq1$ with vanishing boundary condition is considered:
\begin{equation}\label{pde}
\begin{aligned}
\triangle u & =g\\
\left.u\right|_{\partial\Gamma} &= 0
\end{aligned}
\end{equation}
According to \eqref{zamena} the following substitution can be made:
\[
u=v(x_{1},x_{2})\cdot(1-x_{1}^{2}-x_{2}^{2})
\]
where $v$ is to be found using a neural network. To make results comparable to previous cases, analytical solution is chosen as:
\[
u_{a}=f(x_{1},x_{2})\cdot(1-x_{1}^{2}-x_{2}^{2})
\]
where $f(x_{1},x_{2})$ is the expression used for 2D approximation. The function $u_{a}$ is substituted into \eqref{pde} to calculate the source $g$ that would produce it. This way a neural network in the process of minimizing the equation's residual would have to fit exactly the same function as in 2D approximation case. The distinction from direct approximation is that instead of being a squared difference between output and target, $e_{0}$ is a square of the equation's residual. A series of grids is generated with the first one having $\lambda=0.07$ and 704 points and the last one having $\lambda=1.62$ and 3 points. Classical mode is using $s=0$ in \ref{deqloce}, therefore, in addition to values themselves, 4 derivatives of $v$ encountered in the residual $V$ must be calculated in each point:
\[
\frac{\partial}{\partial x_{1}},\frac{\partial}{\partial x_{2}},\frac{\partial^{2}}{\partial x_{1}^{2}},\frac{\partial^{2}}{\partial x_{2}^{2}}
\]
Extended training casts 8 extra operators on $V$:
\[
\frac{\partial}{\partial x_{i}},\frac{\partial^{2}}{\partial x_{i}^{2}},\frac{\partial^{3}}{\partial x_{i}^{3}},\frac{\partial^{4}}{\partial x_{i}^{4}}
\] where $i=1,2$. Due to overlapping this produces only 24 different derivatives of $v$. Therefore, an equivalent classical grid for $M$ points in this mode is $N=5M$. Extended training is run on grids from 139 to 3 points ($\lambda$ from 0.16 to 1.62), and classical one is run on grids from 704 to 16 points ($\lambda$ from 0.07 to 0.6). Test grid has 3000 points and $\lambda=0.033$. Neural networks have the same configurations as for 2D function approximation. Fig. \ref{fig:deq1} compares overfitting ratio between one network trained by extended method and all three processed by classical one. Fig. \ref{fig:deq2} compares performance of the same networks on the test set. Results are similar to those on fig. \ref{fig:2d1} and \ref{fig:2d3}. Overfitting ratios for different architectures trained with extended cost are nearly indistinguishable, so instead, fig. \ref{fig:deq4} depicts the maximum absolute difference between analytical and numerical solution averaged across 5 solving attempts. Extended mode is about one order ahead of classical method on grids that are not too dense.

Particular details for boundary value problem are worth mentioning. As grid spacing $\lambda$ is increased, the last "accurate" solution ($\max|u-u_{a}|<3\cdot10^{-5}$) is obtained by classical method on a grid with $\lambda=0.09$ ($441$ points) and by extended one with $\lambda=0.29$ (52 points). The last "meaningful" solution (${\max|u-u_{a}|<1\cdot10^{-3}}$) is obtained by classical training on a grid with $\lambda=0.17$ (139 points) and by extended one with $\lambda=0.68$ on 12 points. Those results are very close to what one can expect from the 4\textsuperscript{th} and 6\textsuperscript{th} order finite difference stencils. Solution $u$ is based on the equation for $v$, in which terms $4x_{i}\partial v/\partial x_{i}$, $i={1,2}$ would be the biggest source of stencil error $\omega$, and for 2D Poisson equation, the value of $\omega$ is very close to the maximum deviation from the exact solution. The leading term of error of the first order derivative calculated with the 4\textsuperscript{th} order central stencil is $\lambda^{4}/30\cdot \partial^{5} v/\partial x_{i}^{5}$ and with the 6\textsuperscript{th} order stencil is $\lambda^{6}/140\cdot \partial^{7}v/\partial x_{i}^{7}$. After substituting those expressions into the error source and noting that all derivatives of $v$ are of the order of $1$, the 4\textsuperscript{th} order stencil has $\omega=1\cdot10^{-5}$ for $\lambda=0.09$ and $\omega=1\cdot10^{-4}$ for $\lambda=0.17$. The 6\textsuperscript{th} order stencil has $\omega=1.6\cdot10^{-5}$ for $\lambda=0.29$ and $\omega=3\cdot10^{-3}$ for $\lambda=0.68$. However, the 6\textsuperscript{th} order stencil requires at least 7 points in each direction, therefore, it can not be used on a grid with 12 points in total. On the 52-point grid its implementation is possible, but unlikely that precise, since in a lot of points derivatives will have to be calculated using not central but forward or backward stencil, approximation error $\omega$ for which is 20 times higher.

\subsection{Eliminating few trivial explanations}
The simplest explanation of lower overfitting for extended training could be an early stopping. To summarize data the lower 3-quantiles for the number of epochs were calculated for each problem, training mode and network size. For 2D classical approximation they are 7000, 6000 and 4000 and for extended one 5000, 4000 and 4000 for $2\cdot10^{4}$, $1\cdot10^{6}$ and $5\cdot10^{6}$ networks respectively. For 5D case they are 6000, 4000 and 3000 for classical and 6000 for all networks on extended mode. Finally the boundary value problem has got 5000, 3000 and 2000 for classical and 4000 for all networks on extended algorithm. The difference is not quite significant. Even if extended training was run as long as classical whenever possible, it would not bring any noticeable changes to the plots.

Another concern arises from a specific property of RProp. If a landscape of a cost function has a lot of oscillations, then steps for many weights will be reduced to zero quite quickly, and the procedure will not go along those directions. If inclusion of high order derivatives creates such "ripples", training could effectively be constrained to a network of much smaller capacity, therefore, it can exhibit lower overfitting. To investigate this, the following measurement was made: the amount of non zero steps $\delta$ was observed for $2\cdot10^{4}$ network during classical and the 4\textsuperscript{th} order training for 5D function approximation with $N=449$. In both cases the procedure was run for 6000 epochs. It resulted in $\sqrt{\left\langle e_{0}\right\rangle_{\mathrm{train}}}=1\cdot10^{-4}$ and $\sqrt{\left\langle e_{0}\right\rangle_{\mathrm{test}}}=3\cdot10^{-2}$ for classical and $\sqrt{\left\langle e_{0}\right\rangle_{\mathrm{train}}}=2\cdot10^{-3}$, $\sqrt{\left\langle e_{0}\right\rangle_{\mathrm{test}}}=4\cdot10^{-3}$ for extended training. In both cases $\delta\sim300$, except for small intervals of epochs close to the beginning or resetting of steps. The ratio $\delta_{\mathrm{extended}}/\delta_{\mathrm{classical}}$ was calculated with the following results: minimum $0.45$, the lower 3-quantile $0.8$, and median $0.82$. Value $\delta$ is relatively small, so another measurement was made: an average probability of a weight to be changed within 10 epochs is estimated as $11\%$ for classical and $14\%$ for extended training. It does not seem possible that overfitting ratios of 300 and 2 are due to the $\delta$ being 20\% less. Usage of additional derivatives does not restrict training to a smaller subset of weights.

Another simple explanation of overfitting ratio can be obtained as follows: consider 1-dimensional function $f$ approximated by a network $\mathcal{N}$ trained on $n\gg1$ points uniformly distributed on $[0,1]$. Error $e$ on the test set can be approximated as a difference between network's value $\mathcal{N}$ and target $f$ in the middle of an interval between two neighboring training points $a$ and $b$. Its length is $1/n$ and from Taylor expansion in $a$ one can write:
\[
e_{a+\frac{1}{2n}}\sim\left[\mathcal{N}-f\right]_{a+\frac{1}{2n}}\simeq\left[\mathcal{N}-f+(\mathcal{N}-f)'\frac{1}{2n}\right]_{a}
\]
After dividing this expression by $\mathcal{N}-f$ one obtains the overfitting ratio between $e$ on train and test sets:
\begin{equation}\label{ratio1}
\frac{e_{\mathrm{test}}}{e_{\mathrm{train}}}\sim1+\frac{(\mathcal{N}-f)'}{\mathcal{N}-f}\frac{1}{2n}
\end{equation}
If one chooses to split $n$ points into two parts, distribute half of them uniformly and then use the rest to calculate derivatives of target in each point using the first order finite difference stencil, the error in between points and the ratio will be written as:
\[
e_{a+\frac{1}{n}}\simeq\left[\mathcal{N}-f+(\mathcal{N}-f)'\frac{1}{n}\right]_{a}
\]
\begin{equation}\label{ratio2}
\frac{e_{\mathrm{test}}}{e_{\mathrm{train}}}\sim1+\frac{(\mathcal{N}-f)'}{\mathcal{N}-f}\frac{1}{n}
\end{equation}
The term $(\mathcal{N}-f)$ is being minimized by classical algorithm and its derivative $(\mathcal{N}-f)'$ is what is included in the first order extended training. From numerical experiments it was found that, if a neural network trained by classical algorithm shows the train set precision $\varepsilon$ it has the first derivative on this set off by about $10\varepsilon$ (this factor also depends on the network size). For extended mode, the precisions for different trained derivatives are approximately the same $\epsilon$, and the first non trained derivative is similarly $10\epsilon$ off. Therefore, one can estimate test to train ratio \ref{ratio1}:
\[
\frac{e_{\mathrm{test}}}{e_{\mathrm{train}}}\sim1+\frac{10\varepsilon}{\varepsilon}\frac{1}{2n}=1+\frac{5}{n}
\]
And the ratio \ref{ratio2}:
\[
\frac{e_{\mathrm{test}}}{e_{\mathrm{train}}}\sim1+\frac{\epsilon}{\epsilon}\frac{1}{n}=1+\frac{1}{n}
\]
\begin{figure}[!t]
	\centering
	\includegraphics[width=3.5in]{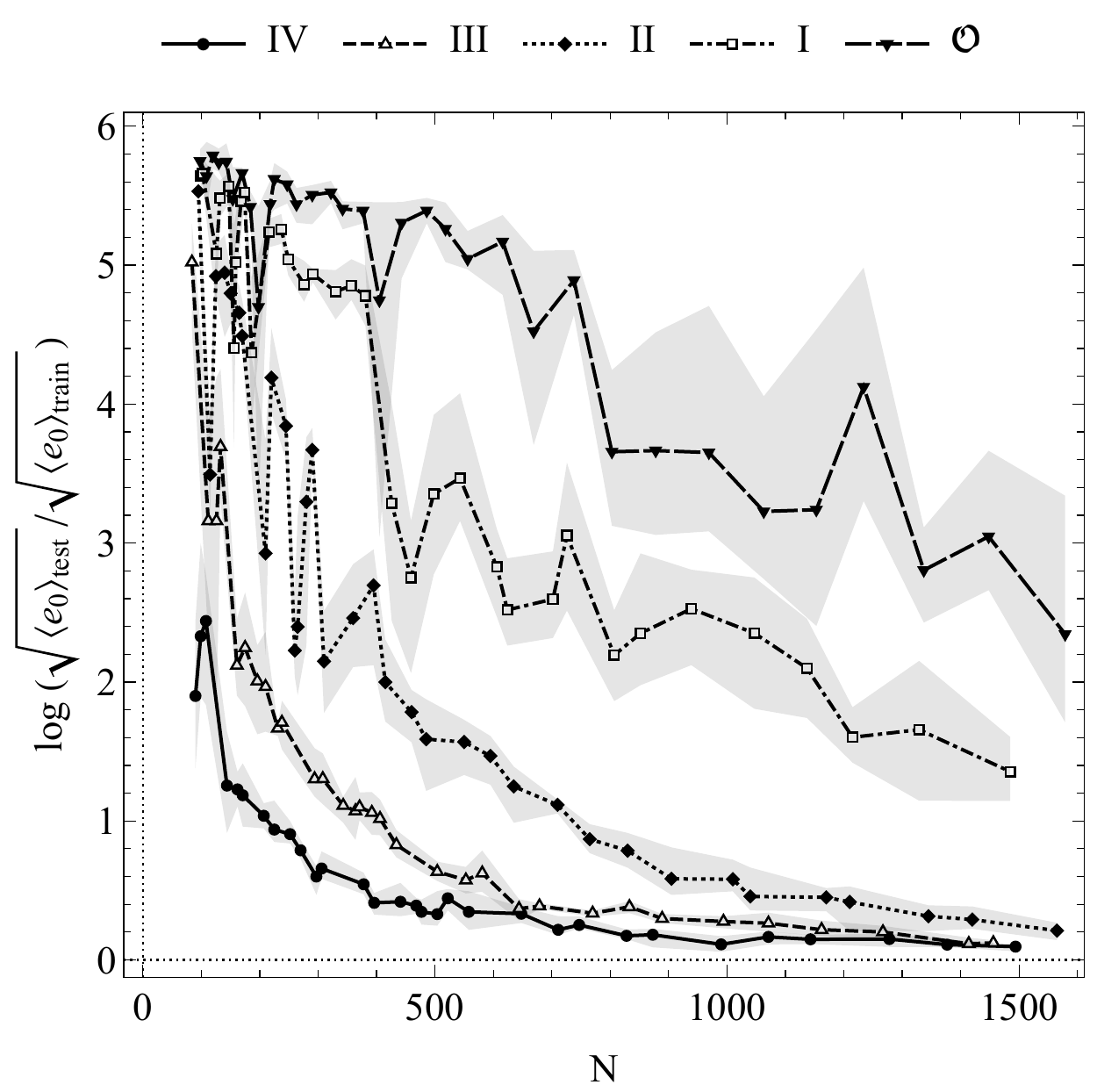}
	\caption{Approximation of 5D function by the network with $10^{6}$ weights. Overfitting ratio lowers as the order of included derivatives increases. Horizontal axis is the number of parameters used for training. Curves are marked according to the maximum order of trained derivatives.}
	\label{fig:varORD}
\end{figure}
For rather small $n$ used in this paper, the difference between logarithms of those expressions is about $0.2$. The same analysis expanded by higher order terms and applied to 2D and 5D cases can more or less explain differences between curves for classical and extended training shown on figure \ref{fig:varORD}, provided ratios $(\mathcal{N}-f)^{\{k\}}/(\mathcal{N}-f)$ are obtained from experiment. Thus, overfitting ratio can be seen as a consequence of two reasons: extended training being simply a higher order approximation and existence of synergy between derivatives, which happens when inclusion of higher order terms enhances precision of the low order ones. However, this description is by no means complete. Note the solid line on fig. \ref{fig:deq1}: from the initial number of 139 grid points down to 11 logarithm of the overfitting ratio for the network with $5\cdot10^{6}$ weights does not exceed $0.03$. Meanwhile, logarithm of the test precision $\log\sqrt{\left\langle e_{0}\right\rangle _{\mathrm{test}}}$ represented by the solid line on fig. \ref{fig:deq2} increases from $-4.5$ to $-2.0$. It means that a network with few millions of connections have not established them in a way that would satisfy 9 conditions in 11 training points more precisely than anywhere else in $\Gamma$. Similar situation can be spotted in 2D and 5D cases of function approximation.
\section{Conclusion}
Unexpectedly low overfitting was observed when derivatives of target up to the 4\textsuperscript{th} order were included in backpropagation. Cost function values on train and test sets were very similar whether neural network approximated a smooth analytical expression or solved partial differential equation. Increasing the capacity of a network from thousands up to millions of weights barely affected those results. Test and train precisions were less than 8\% different for a network with $5\cdot10^{6}$ weights up until 23 points left for 2D function approximation and 11 points left for solving 2D Poisson equation. Whereas approximating smooth analytical functions can be regarded as highly artificial task, solving partial differential equations on a very sparse grids can have more practical applications.


%

\section*{Acknowledgment}
Author expresses a profound gratitude to his scientific advisor E.A.~Dorotheyev and gratefully indebeted to Y.N.~Sviridenko, A.M.~Gaifullin, I.A.~Avrutskaya and I.V.~Avrutskiy without whom this work would not be possible.

\ifCLASSOPTIONcaptionsoff
  \newpage
\fi

\end{document}